\documentclass[10pt,twocolumn,letterpaper]{article}
\usepackage{lineno}
\def\wacvPaperID{237}
\def\confName{WACV}
\def\confYear{2025}

\usepackage{tikz}
\usepackage{booktabs} 
\usepackage{colortbl}
\usepackage{multirow}
\usepackage{subcaption}
\usepackage{tabularx}
\usepackage{caption}
\definecolor{wacvblue}{rgb}{0.21,0.49,0.74}
\usepackage[pagebackref,breaklinks,colorlinks,allcolors=wacvblue]{hyperref}

\usepackage{amsmath, amssymb}
\usepackage[capitalize]{cleveref}
\crefname{section}{Sec.}{Secs.}
\Crefname{section}{Section}{Sections}
\Crefname{table}{Table}{Tables}
\crefname{table}{Tab.}{Tabs.}

\usepackage[pagenumbers]{wacv} 

\definecolor{iccvblue}{rgb}{0.21,0.49,0.74}
\definecolor{CBLightCyan}{HTML}{bae1ff}
\definecolor{CBLightYellow}{HTML}{ffffba}
\definecolor{CBLightGreen}{HTML}{ddf2d1}
\definecolor{LightCyan}{rgb}{0.88,1,1}

\newcommand{\conditioner}{\tau_\theta}

\newcommand{\encoder}{\mathcal{E}}
\newcommand{\decoder}{\mathcal{D}}
\newcommand{\denoiser}{\epsilon_\theta}

\newcommand{\todos}[1]{\textcolor{red}{TODOs}}

\title{Diffusion-Based Action Recognition Generalizes to Untrained Domains}

\author{
Rog\'erio Guimar\~aes\textsuperscript{*} \qquad Frank Xiao\textsuperscript{*} \qquad Pietro Perona \qquad Markus Marks\\
California Institute of Technology\\
{\tt\small rogerio@caltech.edu}
{\tt\small \{rogerio, frank, perona, marks\}@caltech.edu} 
}

\begin{document}
\maketitle

\begingroup
\renewcommand\thefootnote{\fnsymbol{footnote}}
\footnotetext[1]{Equal contribution.}
\endgroup

\begin{abstract}

Humans can recognize the same actions despite large context and viewpoint variations, such as differences between species (walking in spiders vs. horses), viewpoints (egocentric vs. third-person), and contexts (real life vs movies). Current deep learning models struggle with such generalization. We propose using features generated by a Vision Diffusion Model (VDM), aggregated via a transformer, to achieve human-like action recognition across these challenging conditions. We find that generalization is enhanced by the use of a model conditioned on earlier timesteps of the diffusion process to highlight semantic information over pixel level details in the extracted features. We experimentally explore the generalization properties of our approach in classifying actions across animal species, across different viewing angles, and different recording contexts. Our model sets a new state-of-the-art across all three generalization benchmarks, bringing machine action recognition closer to human-like robustness. Project page: \href{https://www.vision.caltech.edu/actiondiff/}{\texttt{vision.caltech.edu/actiondiff}}
Code: \href{https://github.com/frankyaoxiao/ActionDiff}{\texttt{github.com/frankyaoxiao/ActionDiff}}

\end{abstract}

\section{Introduction}
\label{sec:intro}

Analyzing the actions and interactions of agents is potentially one of the most useful applications of computer vision: in autonomous driving~\cite{chang2019argoverse,sun2020scalability}, video gaming~\cite{guss2019minerl, hofmann2019minecraft}, sports analytics\cite{ghosh2023sports}, transportation safety\cite{silva2020machine}, neurobiology~\cite{zelikowsky2018neuropeptide}, ethology~\cite{segalin2021mouse, marks2022deep} and ecology~\cite{mannion2023multicomponent} and myriad other domains. Computer Vision is successful in recognizing agents' actions in large-scale datasets automatically~\cite{jhuang2007biologically, dankert2009automated, branson2009high, mathis2018deeplabcut, segalin2021mouse, sun2023mabe22}. However, successful systems are trained with massive supervision, which requires expensive and tedious labeling, and do not generalize well to new viewing conditions and domains. 
Unsupervised deep learning, which has yielded strong results on many tasks in text and images, has yet to demonstrate similar performance for action recognition in video.

Amongst the success stories of unsupervised learning for image analysis are semantic segmentation \cite{li2023open}, depth estimation \cite{ke2024repurposing, yang2024depth}, drawing 2D correspondences \cite{tang_emergent_2023, luo2023diffusion}, keypoint estimation \cite{khalil2024learning, sun2022self, jakab2020self, zhang2018unsupervised}, and many pixel-level tasks \cite{kondapaneni2024text, zhao2023unleashing} and mid-level vision tasks \cite{chen2024probing}. However, whether the same potential extends beyond the pixel level to a high-level vision task like action recognition is still unclear. 
We are particularly interested in action classification across species~\cite{dell2014automated,datta2019computational} and imaging modalities. 

\begin{figure}
    \centering
    \includegraphics[width=0.475\textwidth]{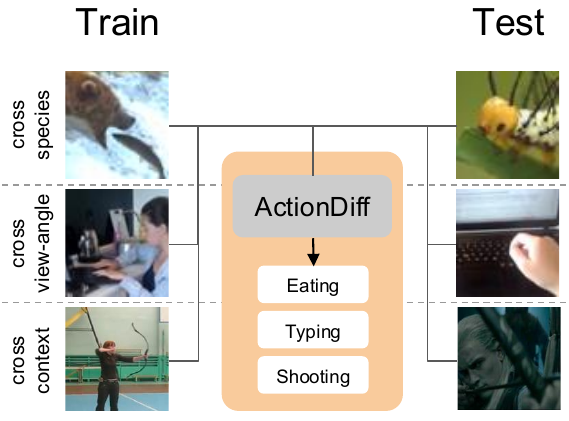}
    \caption{{\bf ActionDiff.} Our method uses the highly semantic features extracted from a frozen Stable Video Diffusion backbone to perform action recognition in tasks that require generalization across different domains. Our model generalizes to new agents (species), view-angles ($1^{st}$ to $3^{rd}$), and contexts (sports vs. movies) that were not present in the training data.}
    \label{fig:teaser}
\end{figure}

Since diffusion models can recognize body parts of animals in images, such as feet and ears, and can generalize such capabilities across species, we are asking if this is also true of actions such as walking or eating? Furthermore, previous literature \cite{park_understanding_2023, choi_perception_2022, daras_multiresolution_2022} shows that earlier steps in the diffusion process focus on low-frequency semantic features, while later steps focus on high-frequency pixel-level details. This indicates that conditioning the diffusion model on these earlier steps can lead to features that are more semantic and robust to generalization. This is in stark contrast to previous works that use diffusion as a backbone for perception tasks \cite{zhao2023unleashing, kondapaneni_text-image_2024}, in which the time conditioning aspect is not explored and models are simply conditioned on the last timestep (lowest noise level).

\begin{figure*}[tp]
    \centering
    \includegraphics[width=\textwidth]{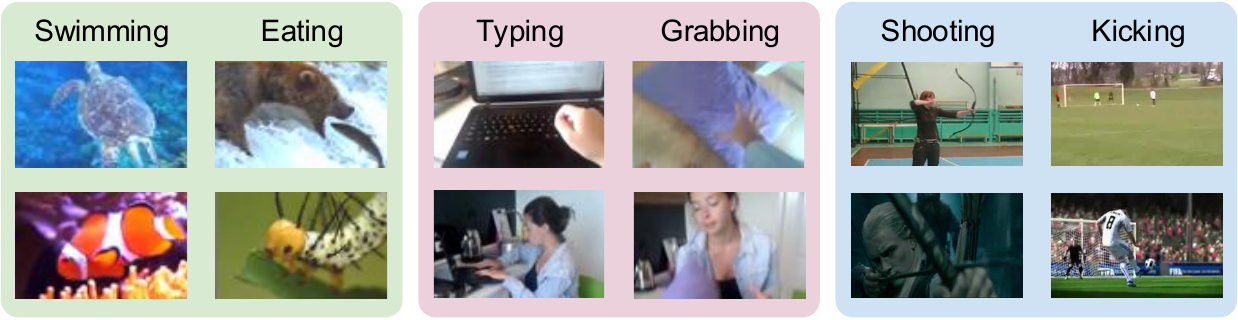}
    \caption{{\bf Domain Shift Tasks.} We present the three domain-shift tasks we use to measure the generalization performance of ActionDiff. {\bf Left:} samples from the Animal Kingdom dataset \cite{ng_animal_2022}, which contains examples of actions (eating and swimming) being performed across different animal species. {\bf Middle:} samples from CharadesEgo \cite{sigurdsson_charades-ego_2018}, which contain examples of the same actions (typing and grabbing a pillow) captured from first and third person perspective. {\bf Right:} Samples from UCF-101 (top) and HMDB51 (bottom), which contain examples of the same actions (shooting a bow and kicking a ball) in different contexts. UCF has mostly amateur sports footage, while HMDB also includes other sources ( such as movies, TV, and video games.)}
    \label{fig:datasets}
\end{figure*}

We explore these theoretical notions by comparing features from video diffusion models conditioned on earlier timesteps across various datasets and tasks.
We first use these features to train a transformer to recognize actions in one animal species and then use the same classifier to recognize the same behavior in another species.
We find that the diffusion model can perform this dramatic semantic shift well, outperforming previously best-performing models.
We then test the same strategy on another critical problem in action recognition: generalizing from third-person to first-person video. We find that diffusion model features are more robust to viewpoint change than the previous state-of-the-art (SOTA).
Lastly, we test on shifts from recognizing actions from different datasets recorded in different contexts, varying video aspects like framing, view angle, lighting, environment, etc.
All in all, we find that our model, based on features of a video diffusion model, handles species, view angle, and context shift better than any of the SOTA models.

Our contributions are as follows:
\begin{itemize}
    \item We introduce \textit{ActionDiff}, a novel action classification method leveraging video diffusion models as a feature extraction backbone, achieving state-of-the-art performance on video perception tasks. 
    \item We systematically benchmark diffusion-derived features across tasks explicitly designed to evaluate robustness against domain shifts in video data to test if the high semantic level of their features helps in generalization. 
    \item We are the first work to analyze the role of time conditioning in diffusion as a backbone for video, and to experimentally show that features learned in earlier steps are more robust to domain shift. Therefore, we find an application for the theoretical prediction that the features learned at such steps focus on lower-frequency signals and are potentially more semantic and general.
\end{itemize}

\section{Related Works}
\label{sec:related_works}

\subsection{Action Recognition}

Action Recognition (AR) is a computer vision task in which a model $f$ is given a sequence of video frames $\mathbf{x}$ and must predict $y=f(\mathbf{x})$, the label of the action(s) present in the video, given a predefined set of possible actions. Early works on AR used spatial-temporal convolutional architectures \cite{carreira_quo_2018, feichtenhofer_spatiotemporal_2017, wang_spatiotemporal_2019}. Recent works have used attention-based models like Transformers \cite{arnab_vivit_2021, yan_multiview_2022, liu_video_2022, tong_videomae_2022, wang_internvideo_2022, wang_videomae_2023} and MAMBA \cite{li_videomamba_2024, fazzari_selective_2025} to efficiently aggregate long-range spatial-temporal information from different frames.

A more challenging scenario in action recognition is handling domain shift \cite{ng_animal_2022, deng_large-scale_2023, sigurdsson_charades-ego_2018}, which occurs when a model trained on a source dataset encounters a significantly different target dataset during evaluation. This could be when labeled data is scarce in the test setting or when raw data is unavailable ({\em e.g.}, when a model will be deployed in an unseen environment). Previous work has tried to address specific domain shifts like different agents (different species performing the same actions) \cite{ng_animal_2022, mondal_actor-agnostic_2023}, viewpoints (third to first person) \cite{lin_egocentric_2022, zhao_learning_2022}, and contexts (sports footage vs other sources like movies) \cite{yao2021videodg, lin_diversifying_2023}. Furthermore, Unsupervised Domain Adaptation (UDA) techniques have been developed to make use of unlabeled data from the target distribution when training the model \cite{chen_temporal_2019, sahoo_contrast_2021} with labeled data from the source distribution, which helps shift from synthetic to real data. ActionDiff, however, addresses all of these domain shift cases without access to target data and using the same model, relying on the highly semantic features of the pretrained video diffusion backbone

\subsection{Video Backbones}
\label{backbones}

Recent works usually employ backbones pre-trained on large-scale datasets that are then fine-tuned on a specific domain. Backbones can be pretrained with supervision in discriminative tasks on large labeled datasets like ImageNet \cite{krizhevsky_imagenet_2012, he_deep_2016, wang_pyramid_2021, liu_swin_2021} or use self-supervised pretraining techniques like contrastive learning \cite{wu_unsupervised_2018, he_momentum_2020, chen_simple_2020, chen_exploring_2020}, or masked modeling \cite{he_masked_2021, bao_beit_2022, xie_simmim_2022}. Recently, a new class of vision-language models like CLIP \cite{radford_learning_2021} have also been used, especially for zero-shot tasks like open vocabulary object detection \cite{zhong_regionclip_2021, wu_cora_2023, wu_clipself_2024}.

Video backbones follow a similar trend but advance slower due to challenges such as scarcity of datasets, extra computation and memory required for processing, and the inherent harder challenge of modeling temporal relations. Most of the work in Action Recognition relies on models pretrained on Kinetics-400 \cite{kay_kinetics_2017}, using CNNs \cite{wang_temporal_2016, lin_tsm_2019, liu_tam_2021, wang_tdn_2021, carreira_quo_2018, tran_closer_2018, wang_appearance-and-relation_2018, feichtenhofer_slowfast_2019} and Transformers \cite{bertasius_is_2021, liu_video_2022, li_uniformer_2022}. More recent works have used masked autoencoding for self-supervised pretraining using transformers \cite{wang_bevt_2022, wei_masked_2023, tong_videomae_2022, wang_videomae_2023, feichtenhofer2022masked}.

When training self-supervised models for image or video with a masked modeling reconstruction loss, models need to predict fine-grained details that do not contribute to the overall semantic understanding of an image, like the exact disposition and contours of leaves on a tree branch. It has been suggested that this decreases the semantic level in the features obtained by such models \cite{assran_masked_2022}. A few techniques have been developed to increase semantic information in the features of generative models like joint embedding \cite{baevski_data2vec_2022, assran_self-supervised_2023, garrido_learning_2024, bardes_revisiting_2024} and self-distillation \cite{caron_emerging_2021, oquab_dinov2_2024}. Still, recent work shows that using features extracted from diffusion backbones can be inherently highly semantic in images \cite{tang_emergent_2023}.

\subsection{Diffusion for Perception Tasks}

The quality of diffusion-generated images \cite{rombach_high-resolution_2022} led to works that tried to use a diffusion model as backbone to extract features from images to be used in downstream tasks like keypoint correspondence, depth estimation, and semantic segmentation \cite{hedlin_unsupervised_2023, tang_emergent_2023, zhao_unleashing_2023, luo_diffusion_2023, hedlin_unsupervised_2024, kondapaneni_text-image_2024, namekata_emerdiff_2024}. However, given the many design choices that can be made with diffusion, the field is not yet set on the optimal method for extracting such features. Some works extract just the intermediate representations of the denoising U-Net \cite{tang_emergent_2023, luo_diffusion_2023}, but some also concatenate the cross-attention maps \cite{zhao_unleashing_2023, kondapaneni_text-image_2024}, while some extract only such maps \cite{hedlin_unsupervised_2023, hedlin_unsupervised_2024}. Regarding the text conditioning, some models use no conditioning \cite{tang_emergent_2023, luo_diffusion_2023}, some use the average embeddings of sentences that are meaningful to the task \cite{zhao_unleashing_2023}, some use automatic captioners \cite{kondapaneni_text-image_2024}, and some optimize the embeddings for the specific task \cite{hedlin_unsupervised_2023, hedlin_unsupervised_2024}. Some approaches add noise to an image before passing it through the denoiser \cite{tang_emergent_2023}, some do not \cite{zhao_unleashing_2023, kondapaneni_text-image_2024}, and some. Questions like what denoiser layer and timestep conditioning should be used have been explored \cite{luo_diffusion_2023} and even whether noise needs to be added to the images before passing them through the denoiser \cite{stracke_cleandift_2024}.

A natural extension of this work is to use diffusion backbones for video tasks. Several works have tried to process video frames independently using a diffusion backbone to extract features for a downstream task \cite{alimohammadi_smite_2025, leonardis_macdiff_2025}. However, a more suitable approach for video processing is to use a video diffusion model as the backbone—a class of generative models that explicitly incorporates a temporal dimension into their latent space to generate coherent, temporally aligned video frames. Early models used a 3D U-Net to process the extra dimension \cite{ho_video_2022}, but most recent works use spatial-temporal attention blocks, like Stable Video Diffusion \cite{blattmann_stable_2023} and ModelScope \cite{wang_modelscope_2023}. Since they became available, some works have used these pretrained models for downstream video tasks like action recognition \cite{weng_genrec_2024}, semantic segmentation \cite{wang_zero-shot_2024}, 3D scene understanding \cite{man_lexicon3d_2024}, and referring video object segmentation \cite{zhu_exploring_2024}.

One of the most highlighted properties of diffusion features is their high semantic content, which emerges naturally in the diffusion pretraining process \cite{tang_emergent_2023, hedlin_unsupervised_2023, namekata_emerdiff_2024}. It has also been shown that this attention to semantics as opposed to fine details increases in the denoiser layers closer to the bottleneck \cite{voynov_p_2023} and in earlier timesteps \cite{park_understanding_2023, choi_perception_2022, daras_multiresolution_2022}. As mentioned in \ref{backbones}, this is a desirable property in pretrained backbones for generalization, but no prior work has taken advantage of these semantic features for this end. Our work shows that the highly semantic features from pretrained Stable Video Diffusion (SVD) outperform SOTA models in tasks that require domain generalization, like adapting to novel agents, viewpoints, and environments.

\begin{figure*}[tp!]
    \centering
    \includegraphics[,width=\textwidth]{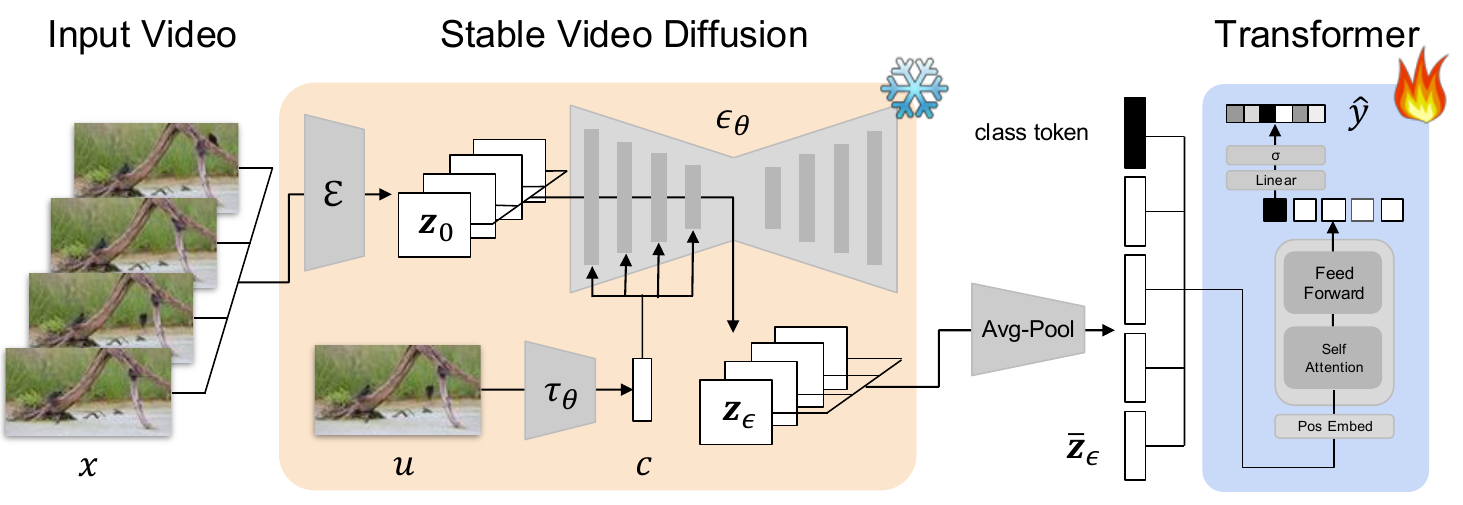}
    \caption{{\bf ActionDiff Architecture.} We split a longer video into shorter segments and extract frame features for each video segment using a frozen Stable Video Diffusion backbone. The video segment frames are encoded into the diffusion latent space by $\mathcal{E}$. They are processed together by the denoiser $\epsilon_\theta$, in a process guided by a condition $c$ (the middle frame $x^{mid}$ embedded by a CLIP encoder $\tau_\theta$) through cross-attention. We extract the outputs of a middle layer $l$ in the denoiser $\epsilon_\theta$, and average pool the outputs across the spatial dimensions to end up with a feature vector for each frame in the video segment. We then collect the sequence of frame features from all video segments and pass them through a learned transformer encoder and a learned class token concatenated to the beginning of the sequence. From the output, we apply a linear layer and a normalization function $\sigma$ to the class token to obtain probabilities $\hat{y}$ for each action class.} 
    \label{fig:architecture}
\end{figure*}

\section{Method}
\label{sec:method}
\subsection{Stable Video Diffusion}

A Latent Diffusion Model (LDM) consists of four networks: An image encoder $\encoder$ , a conditional denoising autoencoder $\denoiser$, a condition encoder $\conditioner$, and an image decoder $\decoder$. The image encoder and decoder are pretrained to convert an image $x_0$ into a latent $z_0$ such that $\encoder(x_0) = z_0$ and $\decoder(z_0)\approx x_0$. The condition encoder is trained to encode some type of conditioning $u$ into a latent $c = \conditioner(u)$ that will guide the diffusion process. The conditional denoising autoencoder is trained to predict the noise $\epsilon \sim \mathcal{N}(0,1)$ added to a latent at a given timestep of a forward diffusion process, given the latent $z_0$, the timestep $t$, and the conditioning $c$, we minimize the loss:
\begin{equation}
L_{LDM} := \mathbb{E}_{\mathcal{E}(x), \tau_\theta(y), \epsilon \sim \mathcal{N}(0, 1), t }\Big[ \Vert \epsilon - \epsilon_\theta(z_{t},t, c) \Vert_{2}^{2}\Big] \, ,
\label{eq:cond_loss}
\end{equation}
Stable Diffusion (SD) \cite{rombach_high-resolution_2022} is a pretrained LDM that uses the VQ-GAN encoder and decoder \cite{esser_taming_2021} for $\encoder$ and $\decoder$, a U-Net \cite{ronneberger_u-net_2015} for $\denoiser$, and a CLIP text encoder \cite{radford_learning_2021} as $\conditioner$ to generate images from text. The U-Net incorporates the conditioning information from $c$ via cross-attention in each layer in the denoising process.

A latent diffusion model can be modified for video generation by introducing temporal attention layers between the network layers used as $\denoiser$ \cite{blattmann_align_2023}. This way, a sequence of frames $\bold{x}$ can have each frame individually encoded through $\encoder$ to generate a latent with an added temporal dimension $\bold{z}$ and denoised by $\denoiser$ through layers that alternate into processing each frame individually and attending to all other frames in the sequence. 

Stable Video Diffusion (SVD) \cite{blattmann_stable_2023} uses this architecture to train a video generator diffusion model trained on a large dataset of paired video and captions curated by them. Their model can generate video from text using the CLIP text encoder as $\conditioner$, but also from a still frame, using the CLIP image encoder instead.

\subsection{Feature Extraction}
\label{extraction}

Let $x^m$ be the $m$-th frame of a video segment of $M$ frames, and let the latent associated with that frame be $z_0^m=\encoder(x^m)$. Let us name the whole sequence of latents be $\mathbf{z}_0=(z_0^1, z_0^2, ..., z_0^M)$. Then we can pass the latent through the denoiser for a single denoising step conditioned on some $t$ and $c$ and we will have $\mathbf{z}_{t,c}=\denoiser(\mathbf{z}_{0},t,c)$ and we will refer to the $m$-th image latent in sequence $\mathbf{z}_{t,c}$ as $z_{t,c}^m$. Notice that after being processed by $\denoiser$ the latents are not independent from each other anymore, since the temporal attention of the denoiser will infuse information from all frames in the sequence in each latent.

Since $\denoiser$ is usually a deep neural network (A U-Net in the case of SVD), let $L$ be its number of layers. We can refer to the output of the $l$-th layer of the denoiser when it's calculating $\denoiser(\mathbf{z}_{0},t,c)$ as $\mathbf{z}_{t,c}^l$ and to the $m$-th latent in this sequence of intermediate latents as $z_{t,c}^{m,l}$. For a predefined set of $t$, $c$, and $l$, we will use $z_{t,c}^{m,l}$ is a latent of dimensions $(h_l,w_l,c_l)$, in which the height, width, and number of channels depend on the layer $l$. Let the sequence of latents $z_{t,c}^{m,l}$ be $z_\epsilon$. We then apply average pooling on the height and width of the feature vector of $c_l$ channels representing the $m$-th frame in our original video segment. We will refer to the average-pooled latents sequence for all video segment frames as $\mathbf{\bar{z}}_\epsilon$.

There are memory constraints on GPUs that will not allow the processing of long videos all at once by $\denoiser$. In that case, we extract the features by grouping frames in windows of length $w$ that fit the GPU and processing each window individually.

\subsection{Action Recognition Head}

Given our sequence of latents $\mathbf{\bar{z}}_\epsilon$, we need a model $f_\phi$ that we will train to predict whether an action is present in the video. As mentioned in \ref{extraction}, longer videos can't have all the features being processed at once by $\epsilon$, and therefore, earlier features are not injected with information from later features that could refine their representation.

To allow for more information mixing, we process the sequence of features using a transformer encoder $f_\phi$ \cite{vaswani_attention_2023}. The transformer receives a sequence of tokens as input, and we will use the average-pooled latents $\mathbf{\bar{z}}_\epsilon$ of our Video Diffusion backbone plus a learned class token \cite{devlin_bert_2019} as the input sequence. In each of its layers, the transformer decoder will perform multi-headed self-attention between the tokens. In this case, the output $f_\phi(\mathbf{\bar{z}}_\epsilon)$ will be a sequence of tokens with the same length as the input sequence $(\mathbf{\bar{z}}_\epsilon+1)$.

To predict the final output $\hat{y}$, the probability that each action is present in the video, we extract the class token from the transformer output and pass it through a linear layer to obtain logits for each action, then through a normalization function $\sigma$ to obtain the final probabilities. When on a multi-label task setting, we use a sigmoid as $\sigma$ and binary cross-entropy loss on each probability. We use softmax as $\sigma$ and cross entropy loss in a single-label task setting. During training, we also use Focal Loss \cite{lin_focal_2018} and Mix-Up \cite{zhang_mixup_2018}.

\begin{table}[tp]
    \centering
    \begin{tabular}{lcc}
        \toprule
        \multirow{2}{*}{Model} & Full           & Unseen \\
              & Dataset (mAP)  & Species (acc) \\
        \midrule
        \multicolumn{3}{l}{\em Published Methods} \\
        \midrule
        CARe \cite{ng_animal_2022}                 & 25.25 & 39.7 \\
        MSQNet \cite{mondal_actor-agnostic_2023}   & 73.10 & 42.5 \\
        Mamba-MSQNet \cite{fazzari_selective_2025} & 74.60 & -    \\
        \midrule
        \multicolumn{3}{l}{\em Self-Supervised Frozen Backbones (our impl.)} \\
        \midrule
        VideoMAEv2 \cite{wang_videomae_2023}       & 63.90 & 37.25 \\
        V-JEPA \cite{bardes_revisiting_2024}       & 78.64 & 51.40 \\
        SDv2 \cite{rombach_high-resolution_2022}   & 78.66 & 41.96 \\
        ActionDiff (ours)                     & {\bf 80.79} & {\bf 51.49} \\
        \bottomrule
    \end{tabular}
    \caption{{\bf Results on Animal Kingdom dataset.} ActionDiff beats the SOTA and other self-supervised frozen backbones in action recognition both on the full dataset and the unseen species partition, in which the species used at test time were not seen during training.}
    \label{tab:animal kingdom}
\end{table}

\section{Experiments}
\label{sec:experiments}

To show the semantic content of the features extracted by SVD, we use it for action recognition in tasks in which domain gaps make it crucial for the model to learn the semantic content of the video. These tasks involve recognizing an action in a different setting than the model was trained on: testing on drastically different agents (different species), view angles (first and third person), and framings (fixed surveillance cameras vs intentional framing on movies and YouTube).

In all of our experiments, in addition to ActionDiff, we report the previously published results and the performance of training our classifier (the transformer decoder) using features obtained by two other frozen video backbones pretrained on self-supervised tasks. These backbones are VideoMAEv2 \cite{wang_videomae_2023} pretrained on their {\em UnlabeledHybrid} dataset and V-JEPA \cite{bardes_revisiting_2024} pretrained on their {\em VideoMix2M} dataset. 
This ensures we can directly compare the quality of the backbone features against each other since all other experimental details are the same across backbones.
Specifically, V-JEPA uses pretraining tasks that aim to learn semantic features to help generalization, which makes it a great comparison for evaluating the semantic level of video diffusion features. 

All experiments use the SVD-XT backbone, which is trained to generate clips of up to 25 frames, so we use an extraction window $w=25$. The commonly used number of generative steps of the backbone is 30, so we condition on timestep $20$. Notice that we index our timesteps in the direction of the generative process (reverse diffusion), so earlier steps mean conditioning on higher noise levels, and later time steps are closer to a clean image. Therefore, timestep $20$ is around 1/3 of the process earlier than the final generative timestep usually used in diffusion-based methods \cite{kondapaneni_text-image_2024, zhao_unleashing_2023} for higher semantic value. We extract features from layer 9, before the last downsampling operation.

We also report Stable Diffusion v2 (SDv2) \cite{rombach_high-resolution_2022} as a frozen pretrained backbone by running SVD but ignoring the temporal layers, using only the s layers from the Image Diffusion backbone that SVD builds upon and processing each frame without temporal context.

We condition the denoiser on timestep $t=20$ and extract the features from the layer that obtained the best result in the experiments with the temporal context. Comparing ActionDiff to SDv2 shows us the contribution of the temporal context encoded in the features by the SVD backbone. For other experimental details an hyperparameters, see \cref{sec:exp-details}

\begin{figure}[tp]
    \centering
    \includegraphics[width=0.5\textwidth]{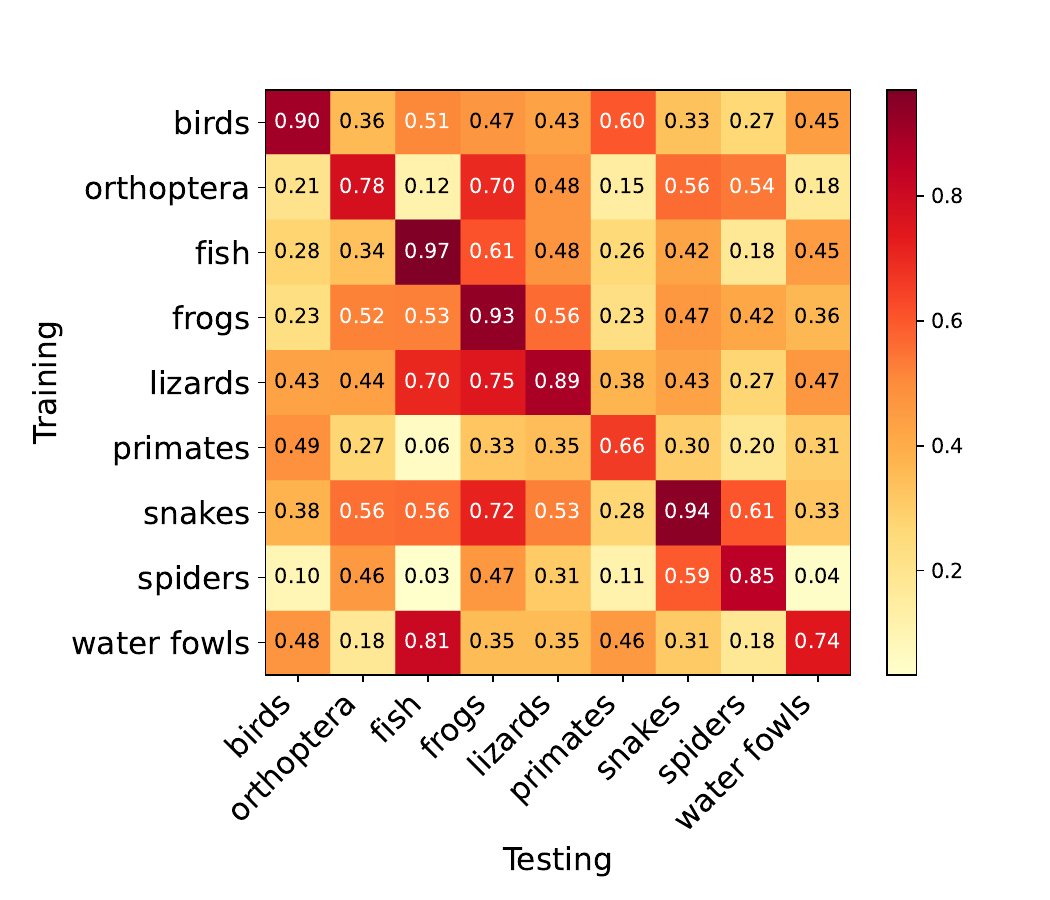}
    \caption{{\bf Species-to-species performance.} We test ActionDiff species-to-species generalization by training and testing in multiple animal type pairs. Results on the main diagonal {\eg birds-birds} show the testing performance in the training data for reference. Overall, we do not find that similarity between species (physical or taxonomic distance helps the model. Conversely, performance is correlated to shared behaviors that cause similar action distributions in the data. For example, a model trained on water fowls (duck and geese) performs much better on fish than on birds.}
    \label{fig:species-to-species}
    \vspace{-0.2in}
\end{figure}

\subsection{Cross-Species Generalization}
The Animal Kingdom dataset \cite{ng_animal_2022} provides 30,100 short videos of 140 actions performed by 650 different animal species. Videos can contain multiple agents, species, and actions. The frequency of an action in this dataset is highly unbalanced across different species. Therefore, to perform well, models must be as close to actor-agnostic as possible to capitalize on training data from actions from one species to another. In \cref{tab:animal kingdom}, we show that ActionDiff outperforms the SOTA model (Mamba-MSQNet) significantly, and SVD features outperform other unsupervised frozen backbones.

Furthermore, we can directly test the cross-species generalization abilities of our model by training and testing on two separate groups of species. To make our results comparable to those reported in \cite{ng_animal_2022}, we use the same set of four animal types for training ({\em i.e.}, birds, fishes, frogs, snakes) and five animal types for testing ({\em i.e.}, lizards, primates, spiders, orthopteran insects, water fowls). We also use only videos labeled with a single action (so we report accuracy instead of mAP). The actions include moving, eating, attending, swimming, sensing, and remaining still. As shown in \cref{tab:animal kingdom}, ActionDiff significantly outperforms the previous state-of-the-art model, MSQNet (50.53 vs. 45.2), as well as other self-supervised frozen backbone methods.

To further analyze what influences our model performance in cross-species generalization, we measured model accuracy when training in a single species and testing on a different single species so that we can compare how well a species transfers to another and assert whether similarity (or proximity in the taxonomic tree) predicts performance (\cref{fig:species-to-species}). Cross-species results correlate more with similar action frequencies between the species than their similarity. For example, training on water fowls works better on fish than in birds (more {\em swimming}), and training on lizards performs better on frogs than snakes (less {\em moving}, more {\em standing still}), which can also be observed by comparing the accuracy matrix (\cref{fig:species-to-species}) with the action frequency correlation matrix (\cref{fig:correlation}) across species. The correlation between accuracy and action frequency correlation is $\rho=0.74$ (\cref{fig:acc_vs_corr}). ActionDiff performance improvements are not due to simply predicting the most likely action, as it significantly outperforms a Na\"ive Bayes approach that predicts the most frequent action in the train set (\cref{fig:gains}), and it beats the other methods in the unseen species task. Action frequencies across species are shown in \cref{fig:freqs}. 

\begin{table}[tp]
    \centering
    \begin{tabular}{lcc}
        \toprule
        \multirow{2}{*}{Model} & $1^{st}\rightarrow1^{st}$ & $3^{rd}\rightarrow1^{st}$\\
        
              & mAP                       & mAP \\
        \midrule
        \multicolumn{3}{l}{\em Fine-tuned Methods (published) } \\
        \midrule
        SSDA \cite{choi_unsupervised_2020}              & 25.8 & 16.6 \\
        AON \cite{sigurdsson_actor_2018}                & 25.9 & -    \\
        ResNet-152 \cite{sigurdsson_charades-ego_2018}  & 27.1 & 19.5 \\
        Ego-Exo \cite{li_ego-exo_2021}                  & 30.1 & -    \\
        EgoVLP \cite{lin_egocentric_2022}               & 32.1 & -    \\
        LaViLa \cite{zhao_learning_2022}                & 36.1 & -    \\
        \midrule
        \multicolumn{3}{l}{\em Self-Supervised Frozen Backbones (our impl.)} \\
        \midrule
        VideoMAEv2 \cite{wang_videomae_2023}            & 16.9      & 15.3 \\
        V-JEPA \cite{bardes_revisiting_2024}            & 23.4      & 17.3 \\
        SDv2 \cite{rombach_high-resolution_2022}        & 35.0      & 29.4 \\
        ActionDiff (ours)                               & \bf{36.5} &  {\bf 30.2} \\
        \bottomrule
    \end{tabular}
    \caption{{\bf Results on the Charades-Ego dataset.} ActionDiff beats the SOTA and other self-supervised frozen backbones in action recognition on $1^{st}$ to $1^{st}$ person and $3^{rd}$ to $1^{st}$ person viewpoint.}
    \label{tab:charades-ego}
\end{table}

\begin{table}[tp]
    \centering
    \begin{tabular}{lcc}
        \toprule
        \multirow{2}{*}{Model} & U $\rightarrow$ H & H $\rightarrow$ U \\
              & Accuracy                   & Accuracy                  \\
        \midrule
        \multicolumn{3}{l}{\em Published Methods} \\
        \midrule
        ADA\cite{volpi2018generalizing} & 56.9 & 72.2 \\
        M-ADA\cite{qiao_learning_2020} & 55.6 & 71.5 \\
        Jigsaw\cite{carlucci_domain_2019} & 55.2 & 72.4 \\
        VideoDG \cite{yao2021videodg}      & 59.1 & 74.9     \\
        STDN \cite{lin_diversifying_2023}                & 60.2       & 77.1 \\
        \midrule
        \multicolumn{3}{l}{\em Self-Supervised Frozen Backbones (our impl.)} \\
        \midrule
        VideoMAEv2 \cite{wang_videomae_2023}      & 23.8 & 26.7     \\
        V-JEPA \cite{bardes_revisiting_2024}      & 51.75       & 60.58 \\
        SDv2 \cite{rombach_high-resolution_2022}  & 76.5  &  79.5  \\
        ActionDiff (ours)                    & {\bf 77.6} & {\bf 81.5} \\
        \bottomrule
    \end{tabular}
    \caption{{\bf Results on the UCF101 (U) to HMDB51 (H) domain shift and vice-versa.} ActionDiff beats the previous SOTA and other self-supervised frozen backbones when trained on any of the datasets and tested on the other. Results for ADA, M-ADA, and Jigsaw were obtained by \cite{yao2021videodg}.}
    \label{tab:UCF-HMDB}
    \vspace{-0.1in}
\end{table}

\begin{figure*}[t!]  
    \centering
    \begin{subfigure}{0.24\textwidth}
        \centering
        \includegraphics[width=\textwidth]{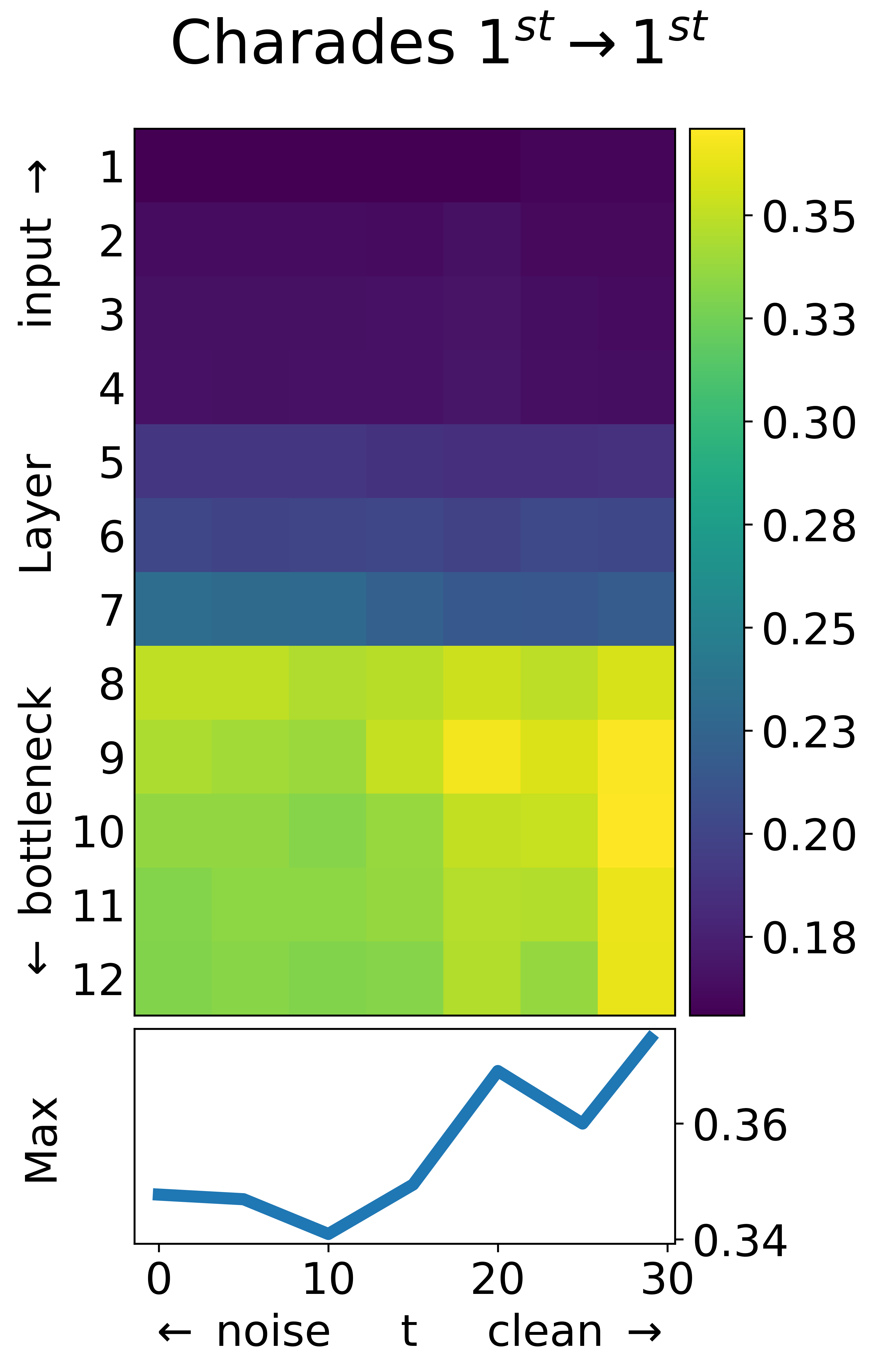}
        \caption{}
    \end{subfigure}
    \begin{subfigure}{0.24\textwidth}
        \centering
        \includegraphics[width=\textwidth]{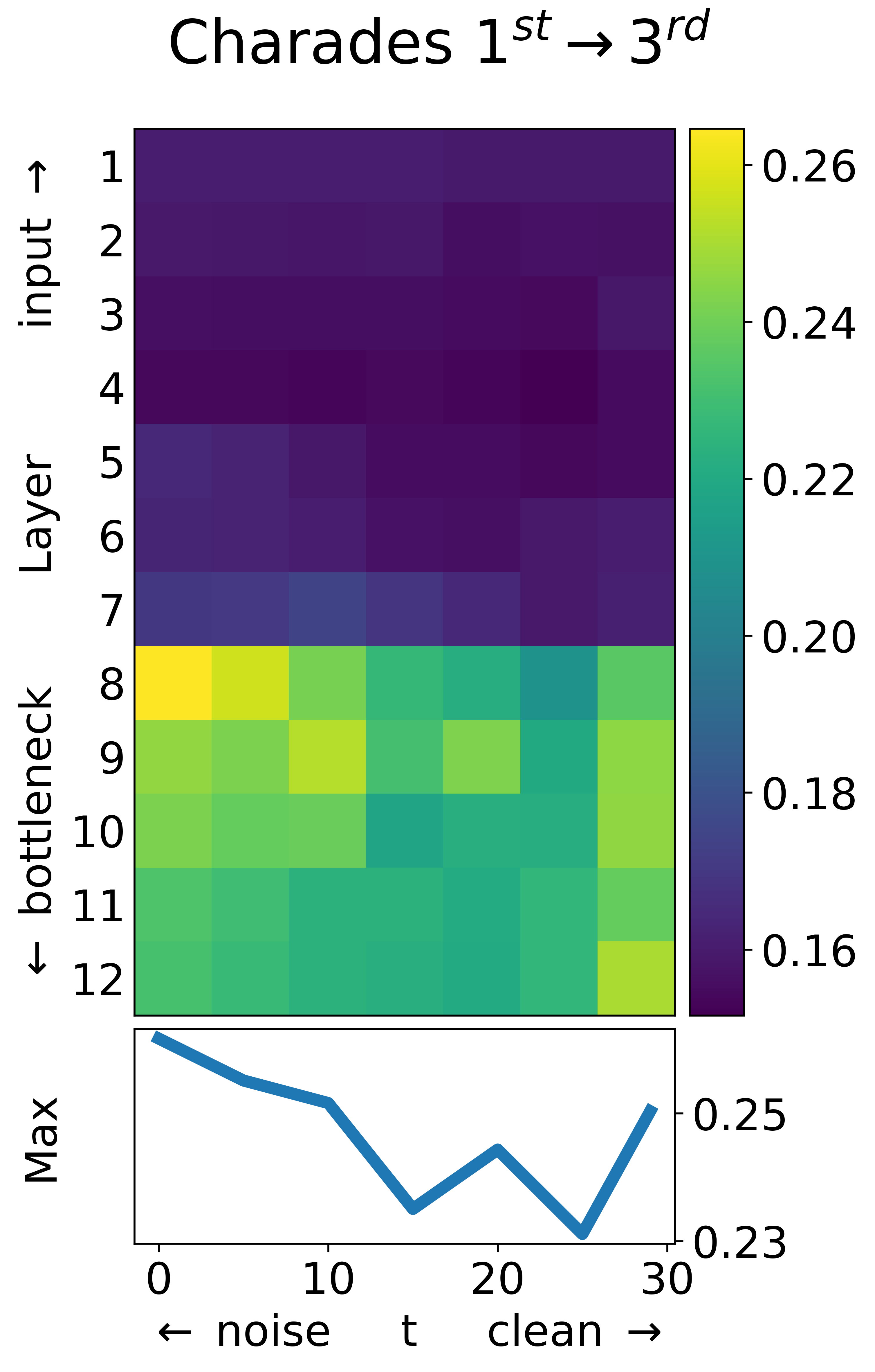}
        \caption{}
    \end{subfigure}
    \begin{subfigure}{0.24\textwidth}
        \centering
        \includegraphics[width=\textwidth]{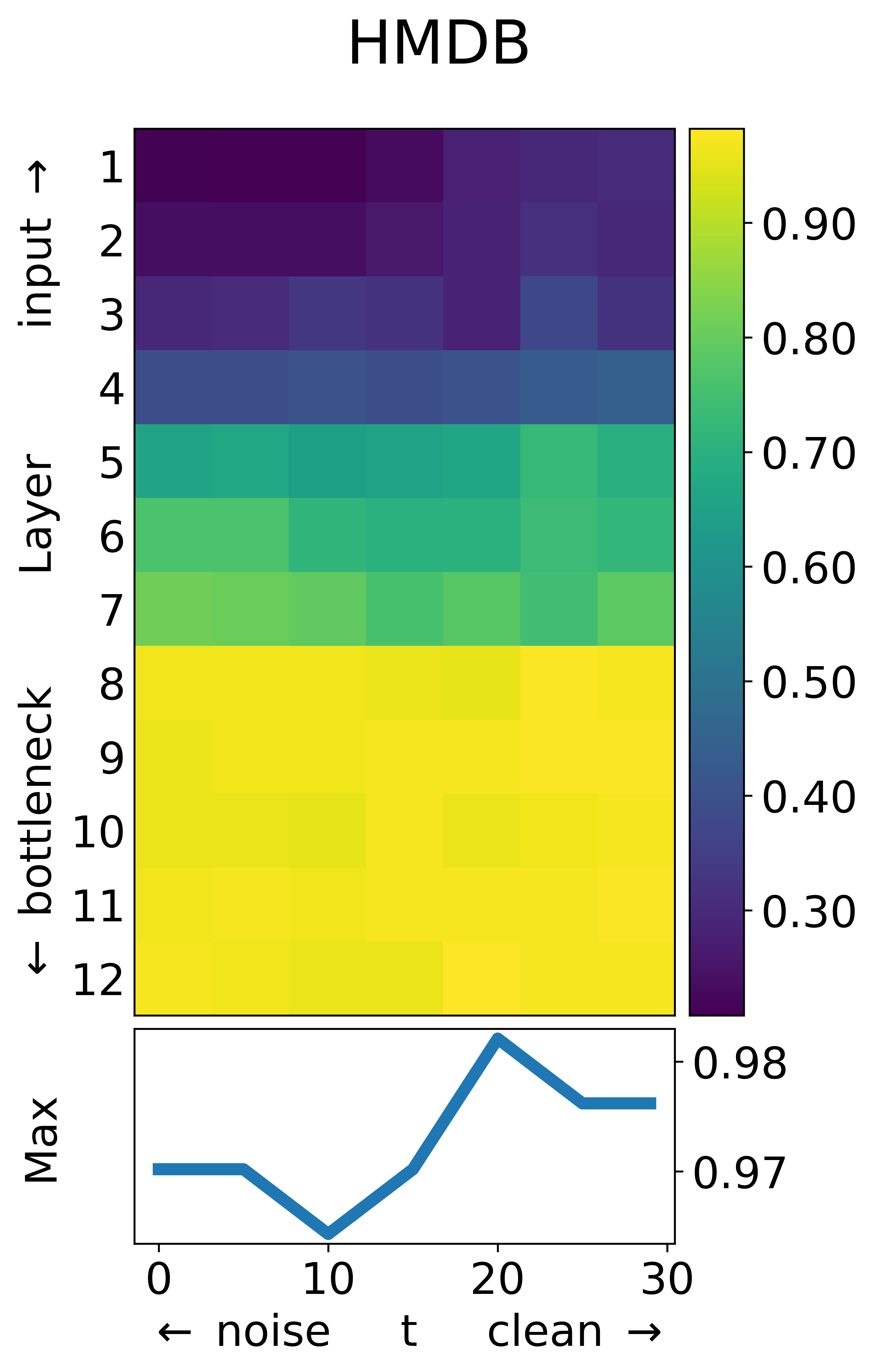}
        \caption{}
    \end{subfigure}
    \begin{subfigure}{0.24\textwidth}
        \centering
        \includegraphics[width=\textwidth]{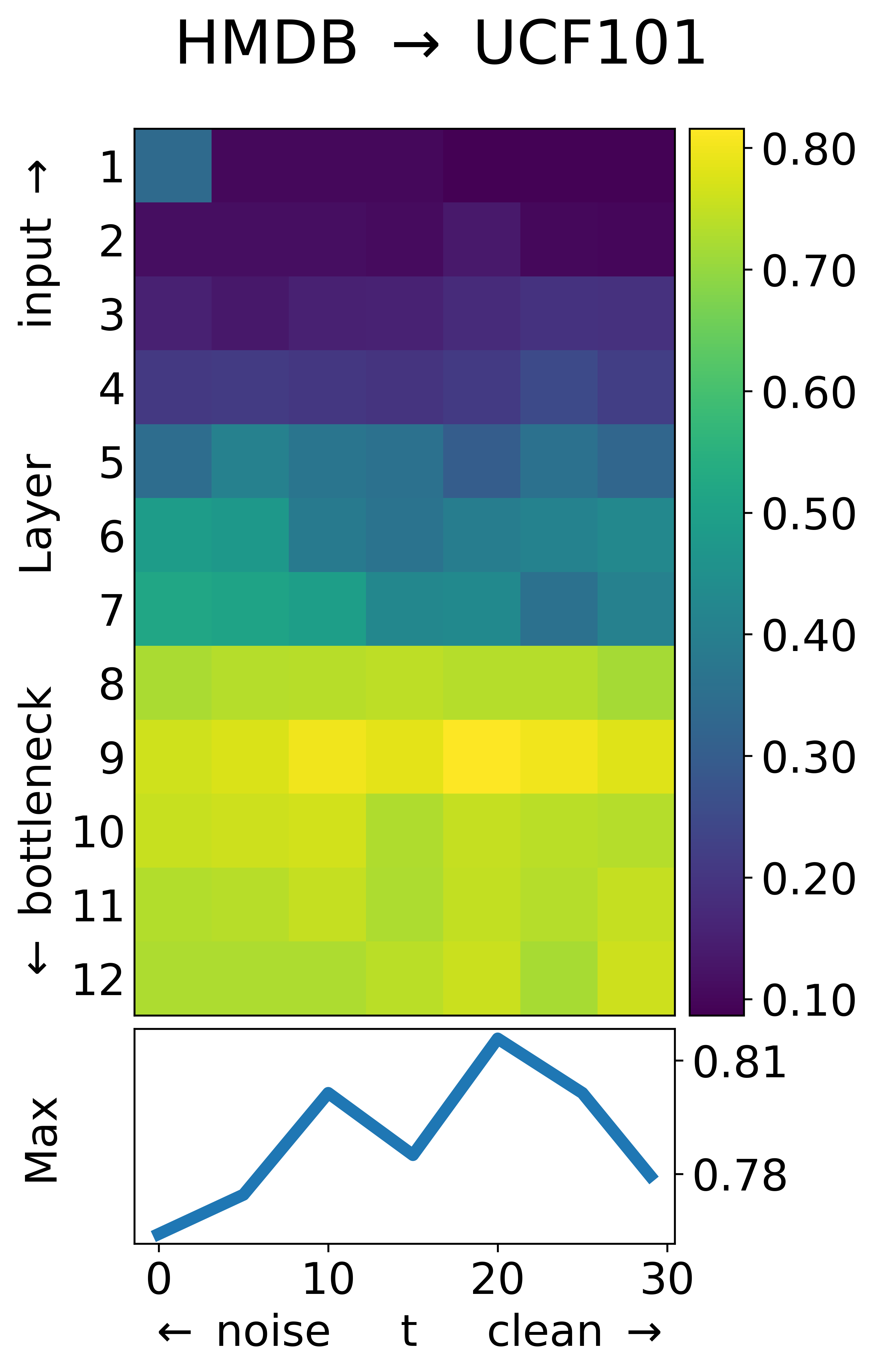}
        \caption{}
    \end{subfigure}

    \caption{{\bf Diffusion Layer and Timestep Conditioning.} We test the performance of our model with diffusion features extracted from different layers and conditioned on different timesteps to analyze the difference between the best features for in- and out-of-domain tasks. We use two example tasks where the model can train on the same train set for in and out-of-domain tasks. Each heatmap shows the results (mAP for CharadesEgo and acc for HMDB to UCF) for each layer (y-axis) and timestep (x-axis) on a task, and the plot below each heatmap shows the best result across layers obtained for each timestep. When we analyze which features are best for in-domain vs out-of-domain testing, we see a shift toward those obtained with earlier timesteps when we test out-of-domain. Layers are indexed in the direction from the input to the bottleneck, and timesteps are indexed in the direction of the generative process (reverse diffusion).}
    \label{fig:gridsearch}
\end{figure*}

\subsection{Cross-Viewpoint Generalization}
\label{sec:viewpoint}
The Charades-Ego dataset~\cite{sigurdsson_charades-ego_2018} consists of 68.8 hours of real-world video footage containing 68,536 annotated action instances captured simultaneously from paired first-person and third-person perspectives.
As shown in~\cref{tab:charades-ego}, our ActionDiff model establishes a new state-of-the-art when trained solely on first-person videos, despite utilizing an SVD backbone that was not fine-tuned specifically on egocentric data. This result is particularly noteworthy since the previous state-of-the-art, LaViLa~\cite{zhao_learning_2022}, relied extensively on pretraining with Ego4D~\cite{grauman_ego4d_2022}, one of the largest egocentric video datasets available.
Furthermore, we also show how the semantic knowledge of diffusion can be used to generalize across viewpoints since our model also beats SOTA results when training the classifier on third-person views and testing on first-person videos. 

\subsection{Cross-Context Generalization}

UCF101 \cite{soomro2012ucf101} is a dataset containing 101 classes of user-uploaded human actions. The HMDB51 \cite{kuehne2011hmdb} is also a human action detection dataset over 51 classes with videos sourced from movies and YouTube videos. In our experiments, we use videos that contain one of the twelve action classes shared between the two datasets, as assembled by \cite{chen_temporal_2019}. The videos from UCF101 are user uploaded and captured in often similar contexts, mostly footage from amateur sports. The videos from HMDB51 are sourced from more unconstrained contexts (such as TV, movies, and video games) that vary environment, lighting, camera framing and angle, as shown in \cref{fig:datasets}. 

\cref{tab:UCF-HMDB} shows our model beats the previous SOTA when adapting from UCF101 to HMDB51, which was obtained by a video domain generalization method \cite{lin_diversifying_2023}, and the other self-supervised frozen backbones we tested. 

\subsection{Diffusion Layer and Timestep Conditioning}

For analysis, we measure the effect of layer and timestep conditioning on the features extracted by the diffusion backbone by grid-searching across all layers $l$ and multiple timesteps $t$. Results in \cref{fig:gridsearch} show the results of evaluating those features in two pairs of in-domain (Charades Ego $1^{st}$ to $1^{st}$ person (a) and HMDB51 (b)) and out-of-domain (Charades Ego $1^{st}$ to $3^{rd}$ person (c) and HMDB51 to UCF101 (d)). When we compare the testing in-domain vs out-of-domain ((a) to (b) and (c) to (d)), we see that earlier steps perform better than the last steps out of domain, a trend not observed in-domain, where the last steps are slightly better. This corroborates previous work showing that earlier diffusion steps focus more on coarse shapes and low-frequency signals, while later steps focus more on fine details and high-frequency signals \cite{park_understanding_2023, choi_perception_2022, daras_multiresolution_2022}.

Our findings indicate optimal performance at earlier timesteps than reported in previous studies, which identified timesteps around 80\% to 90\% of the diffusion process as optimal for keypoint correspondence in images \cite{luo_diffusion_2023}. This difference suggests that earlier timesteps are particularly beneficial for enhancing domain generalization in out-of-domain tasks. Heatmaps (a) and (c) show that timestep conditioning play a very minor role in in-domain tasks across the same layer. It is only in heatmaps (b) and (d), out of domain, that timestep makes a difference within the same layer. This is potentially why effects of timestep conditioning have not been previously reported.

Different tasks can vary in the location of the optimal UNet layer, but all tasks show a trend of better performance in deeper layers. In \cref{tab:layers} we also verify that in all the tasks reported in our experiments, the deeper layer we are using (9) performs better than earlier layers (3 and 6).

\subsection{Ablations}

To test the importance of different parts of our full model, we conduct ablations on the Animal Kingdom dataset with respect to the classifier, extraction window length, and diffusion conditioning. Results are reported in \cref{tab:ablations} and were performed using an SVD backbone pretrained for generating smaller videos (14 frames) than the SVD-XT backbone we use in the other experiments.

In the first part of \cref{tab:ablations}, we see that the transformer encoder has great gains compared to a simple linear or MLP classifier. This shows the importance of the attention mechanism to aggregate the information from the multiple latents representing all the input video frames. In the linear and MLP setting, these frames are concatenated.

Then, we analyze the effect of using different extraction window lengths, which is the number of frames encoded together by the backbone and provide all the temporal context that video diffusion can encode in the features. Throughout all the experiments in this paper, SVD outperforms SDv2, which shows that the video backbone's temporal context provides an advantage compared to the image backbone. However, this advantage saturates quickly with the number of frames in the extraction window, and we see no significant difference in using 10, 16, or 21 frames. 

Next, we measure the effect of two training techniques we used to address train imbalance and help generalization. Due to class imbalance in Animal Kingdom among actions, focal loss \cite{lin_focal_2018} provides gains in performance to our model. Mix-Up \cite{zhang_mixup_2018} also helps since this dataset requires generalization across agents, and thus, memorization is hurtful.

Finally, we tested different conditioning schemes during extraction using our video diffusion backbone. By default, SVD uses a single frame to condition its video generation process, and the results in our previous experiments use the OpenCLIP image embedding \cite{cherti_reproducible_2023} of the middle frame in a window as the conditioning during feature extraction. This method is better than unconditional extraction or conditioning on the OpenCLIP text embedding of the names of all the possible actions we want to recognize in a dataset.

\begin{table}[tp]
    \centering
    \begin{tabular}{lcccc}
        \toprule
        Classifier & Window & Loss & Cond & mAP \\
        \midrule
        Linear                   & 10    & Focal & Frame & 73.0 \\
        MLP                      & 10    & Focal & Frame & 75.2 \\
        Transformer              & 10    & Focal & Frame & 80.8 \\
        \midrule
        Transformer              & 10    & Focal & Frame & 80.8 \\
        Transformer              & 16    & Focal & Frame & 80.9 \\
        Transformer              & 21    & Focal & Frame & 80.8 \\
        \midrule
        Transformer              & 10    & BCE   & Frame & 79.9 \\
        Transformer              & 10    & Focal & Frame & 80.7 \\
        \midrule
        Transformer              & 10    & Focal & None   & 77.4 \\
        Transformer              & 10    & Focal & Action & 74.0 \\
        Transformer              & 10    & Focal & Frame  & 80.8 \\
        \bottomrule
    \end{tabular}
    \caption{{\bf Ablation Studies.} We ablate different parts of our model and vary hyperparameters on the Animal Kingdom Full dataset to show their impact on performance. We test different classifier architectures, extraction window lengths for the backbone, the role of focal loss in an unbalanced dataset, and different conditioning choices available to the Stable Video Diffusion backbone.}
    \label{tab:ablations}
    \vspace{-0.15in}
\end{table}
\section{Discussion}
\label{sec:discussion}

We present ActionDiff, a novel method that utilizes features from video diffusion models for action recognition.
To achieve this, we explore the best way to extract and integrate features from a video diffusion model over time.
We identify the optimal diffusion timestep and layer to extract video diffusion features for action recognition.
We then use these features as input to a transformer.
We use this architecture to test performance on some major action recognition challenges.
We first test on \textit{Cross-Species Generalization}, i.e., how well actions can be recognized across different species, using the Animal Kingdom dataset.
We find that video diffusion features are exceptionally suited for this task, and our approach sets the current state-of-the-art for this challenging task.
Next, using the Charades-Ego dataset, we test how well our approach copes with \textit{View Angle Shifts}.
Again, features from diffusion models perform very well, beating SOTA methods.
Finally, using the UCF-HMDB dataset, we show that the early features from diffusion models also generalize well, identifying the same actions being performed in different contexts, beating SOTA methods available.
Overall, our method sets the current SOTA in multiple action recognition tasks, thereby showing the utility of using diffusion model features.
It is noteworthy that, in the Animal Kingdom dataset, V-JEPA closely matches our results despite being built upon fundamentally different design principles, highlighting the potential of diverse approaches in representation learning.
A notable limitation of our approach is its substantial computational cost, primarily due to the diffusion-based feature extraction, which may hinder scalability and practical deployment in resource-constrained scenarios.
Future research could explore diffusion-based features for other video-related tasks beyond the scope of our current experiments, such as video grounding, temporal localization, or activity recognition. Additionally, integrating parameter-efficient fine-tuning techniques, such as Low-Rank Adaptation (LoRA), to reduce computational overhead represents an exciting direction for making our model more accessible and efficient in practical scenarios.
{
    \small
    \bibliographystyle{ieee_fullname}
    \bibliography{main}
}

\newcommand{\mathbbm}[1]{\text{\usefont{U}{bbm}{m}{n}#1}}
\newcommand{\appendixhead}%

\counterwithout{figure}{section}
\counterwithout{table}{section}
\counterwithout{equation}{section}

\renewcommand\thefigure{S\arabic{figure}}
\renewcommand\thetable{S\arabic{table}}
\renewcommand\theequation{S\arabic{equation}}

\clearpage
\appendix

\section{Supplement}
\label{sec:supplement}

\subsection{Experimental Details}
\label{sec:exp-details}

\subsubsection{Animal Kingdom Experiments}
We employ a transformer encoder with 6 layers and 8 attention heads per layer for all feature representations. The network is trained using a cosine annealing schedule with a peak learning rate of $3 \times 10^{-5}$ over $15$ epochs. We use focal loss with $\alpha=0.25$ and $\gamma=2.0$, apply weight decay of $1 \times 10^{-4}$ and mixup augmentation with $\alpha$ value of $0.2$. 

\begin{figure}[h!]
    \centering
    \includegraphics[width=0.4\textwidth]{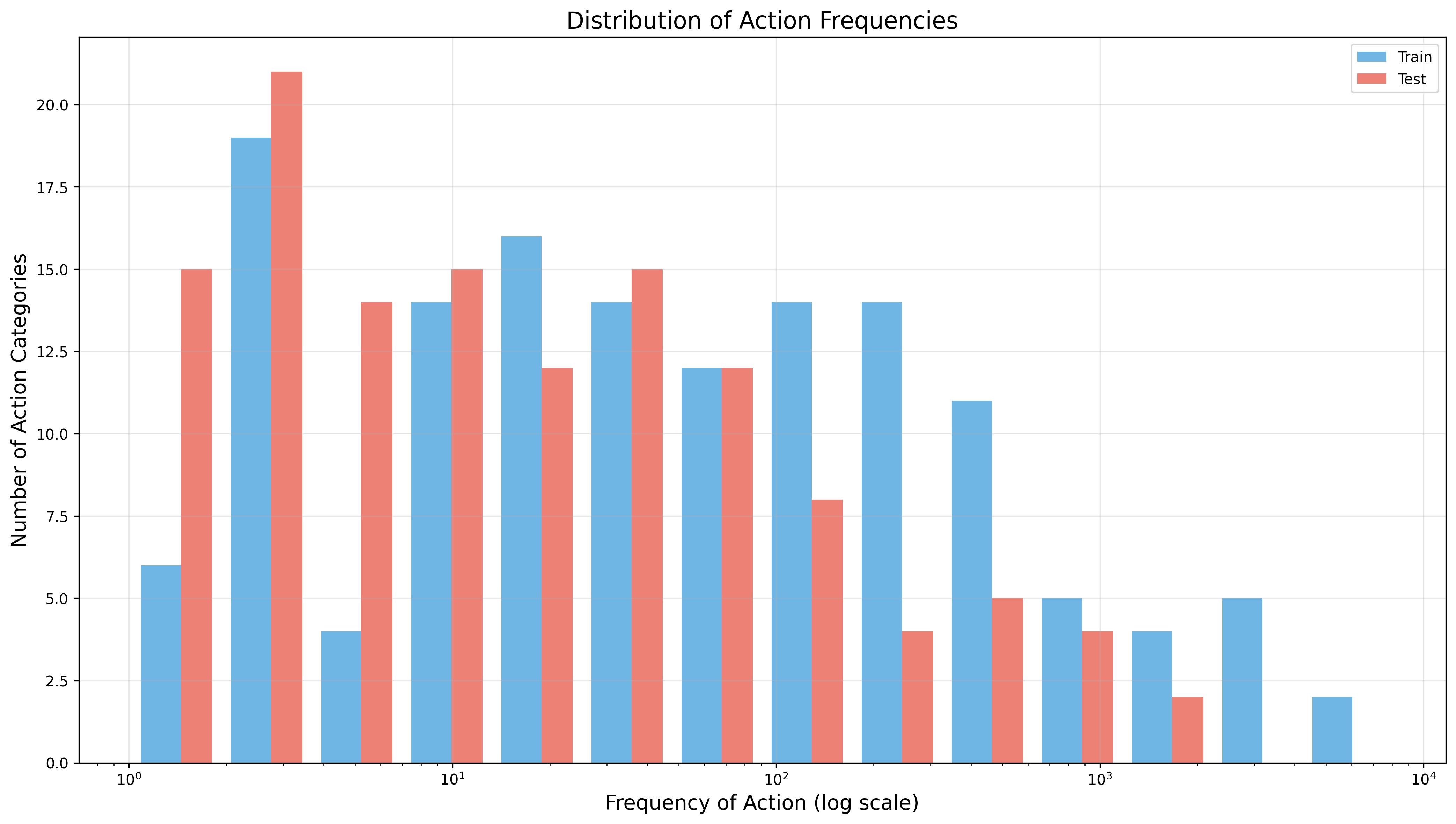}
    \caption{\textbf{Animal Kingdom Action Distribution.}}
    \label{fig:AKActionDistribution}
\end{figure}

\begin{figure}[h!]
    \centering
    \includegraphics[width=0.4\textwidth]{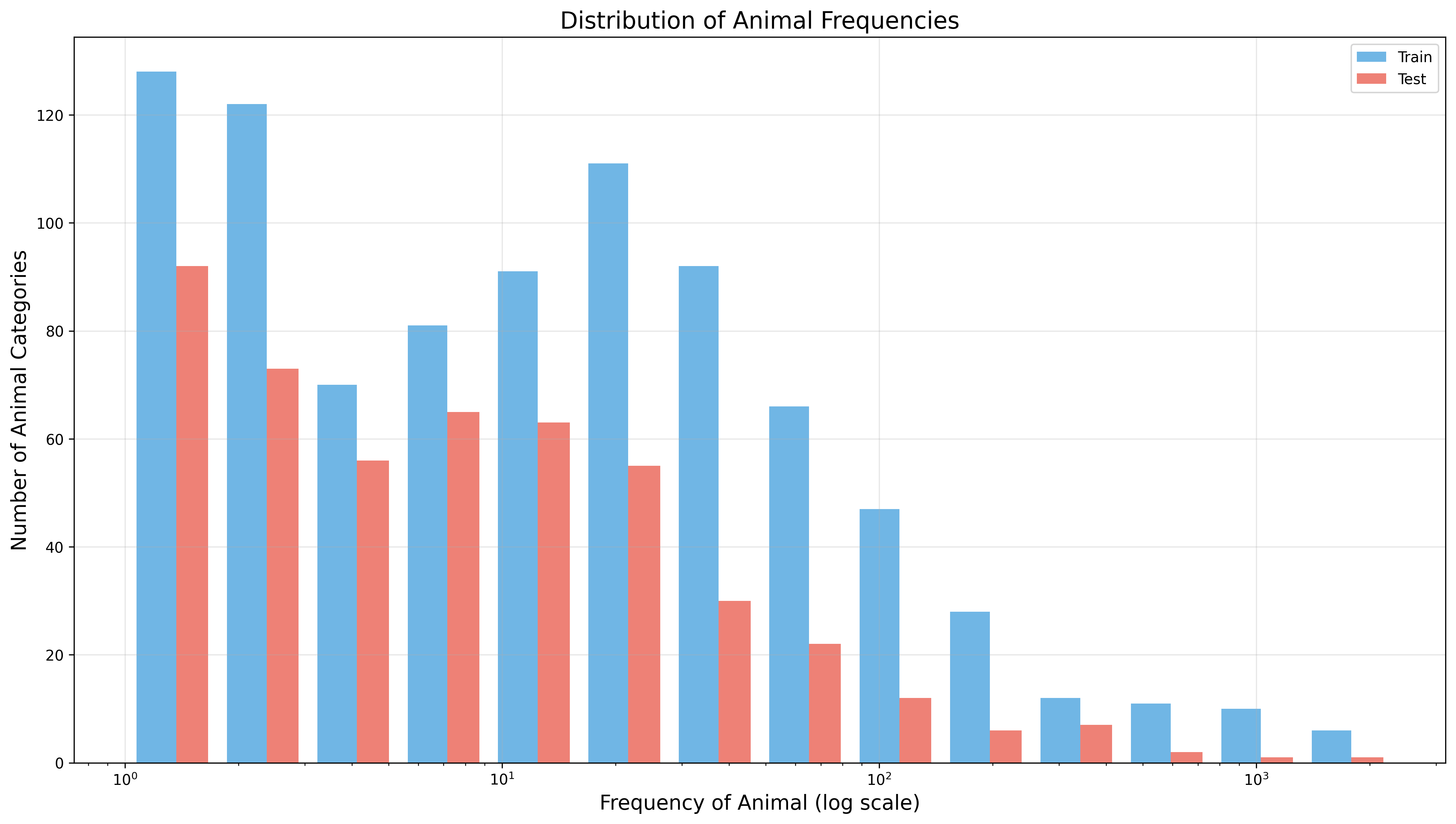}
    \caption{\textbf{Animal Kingdom Species Distribution.}}
    \label{fig:AKSpeciesDistribution}
\end{figure}

The distribution of actions and species are shown in \ref{fig:AKActionDistribution} and \ref{fig:AKSpeciesDistribution}

\subsubsection{CharadesEgo Experiments}
Similarly, transformer encoder with 6 layers and 8 attention heads per layer for all feature representations. The network is trained using a cosine annealing schedule with a peak learning rate of $1 \times 10^{-4}$ over $15$ epochs. We use focal loss with $\alpha=1.0$ and $\gamma=2.0$, apply weight decay of $1 \times 10^{-3}$ and mixup augmentation with $\alpha$ value of $0.3$. 

\begin{figure}[h!]
    \centering
    \includegraphics[width=0.4\textwidth]{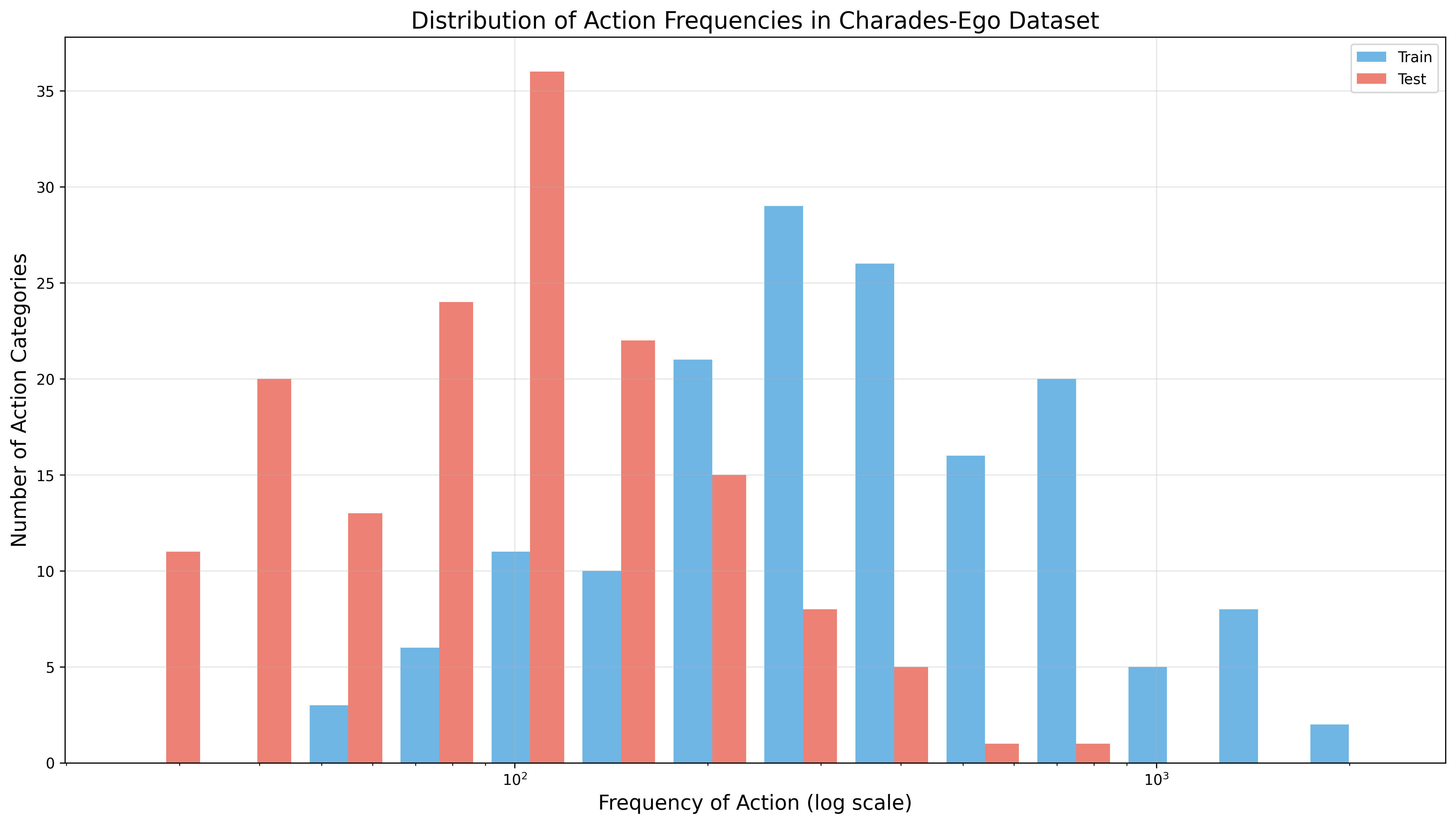}
    \caption{\textbf{Charades-Ego Action Distribution.}}
    \label{fig:CharadesDistribution}
\end{figure}

The distribution of actions is shown in \ref{fig:CharadesDistribution}

\subsubsection{UCF-HMDB Experiments}
We use a transformer encoder with 6 layers and 8 attention heads per layer for all feature representations. The network is trained using a cosine annealing schedule with a peak learning rate of $1 \times 10^{-4}$ over $15$ epochs. We use focal loss with $\alpha=1.0$ and $\gamma=2.0$, apply weight decay of $1 \times 10^{-4}$ and mixup augmentation with $\alpha$ value of $0.0$. 

\subsubsection{Grid-Search}
We use a transformer encoder with 6 layers and 8 attention heads per layer for all feature representations. The network is trained using a cosine annealing schedule with a peak learning rate of $1 \times 10^{-4}$ over $15$ epochs. We use focal loss with $\alpha=1.0$ and $\gamma=2.0$. For CharadesEgo (\cref{fig:gridsearch} a and b), we apply weight decay of $1 \times 10^{-3}$ and mixup augmentation with $\alpha$ value of $0.3$. For HMDB and HMDB-UCF (\cref{fig:gridsearch} c and d), we apply weight decay of $1 \times 10^{-4}$ and mixup augmentation with $\alpha$ value of $0.0$. 

\begin{figure}[h!]
    \centering
    \includegraphics[width=0.4\textwidth]{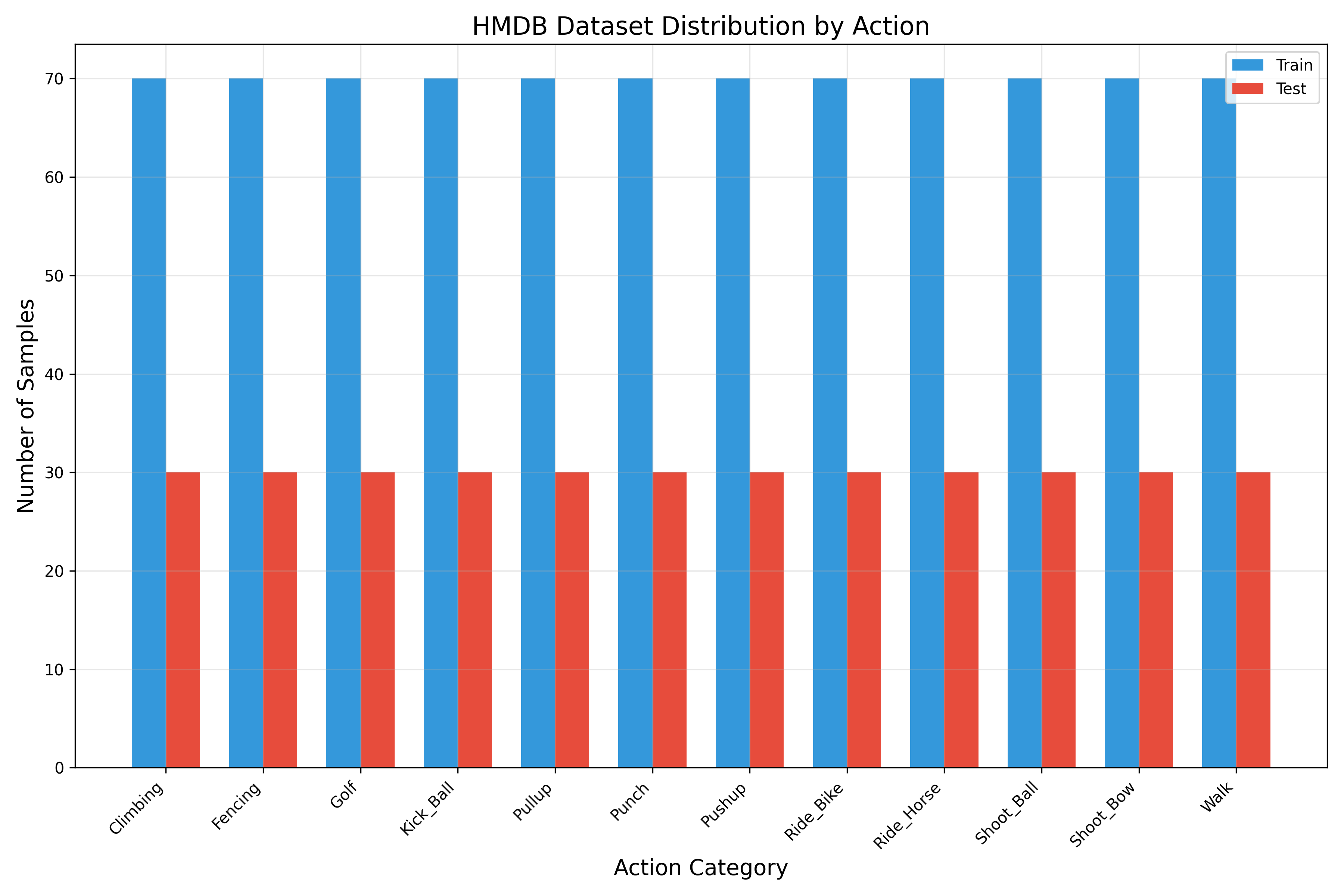}
    \caption{\textbf{HMDB51 Distribution.}}
    \label{fig:HMDBSpeciesDistribution}
\end{figure}

\begin{figure}[h!]
    \centering
    \includegraphics[width=0.4\textwidth]{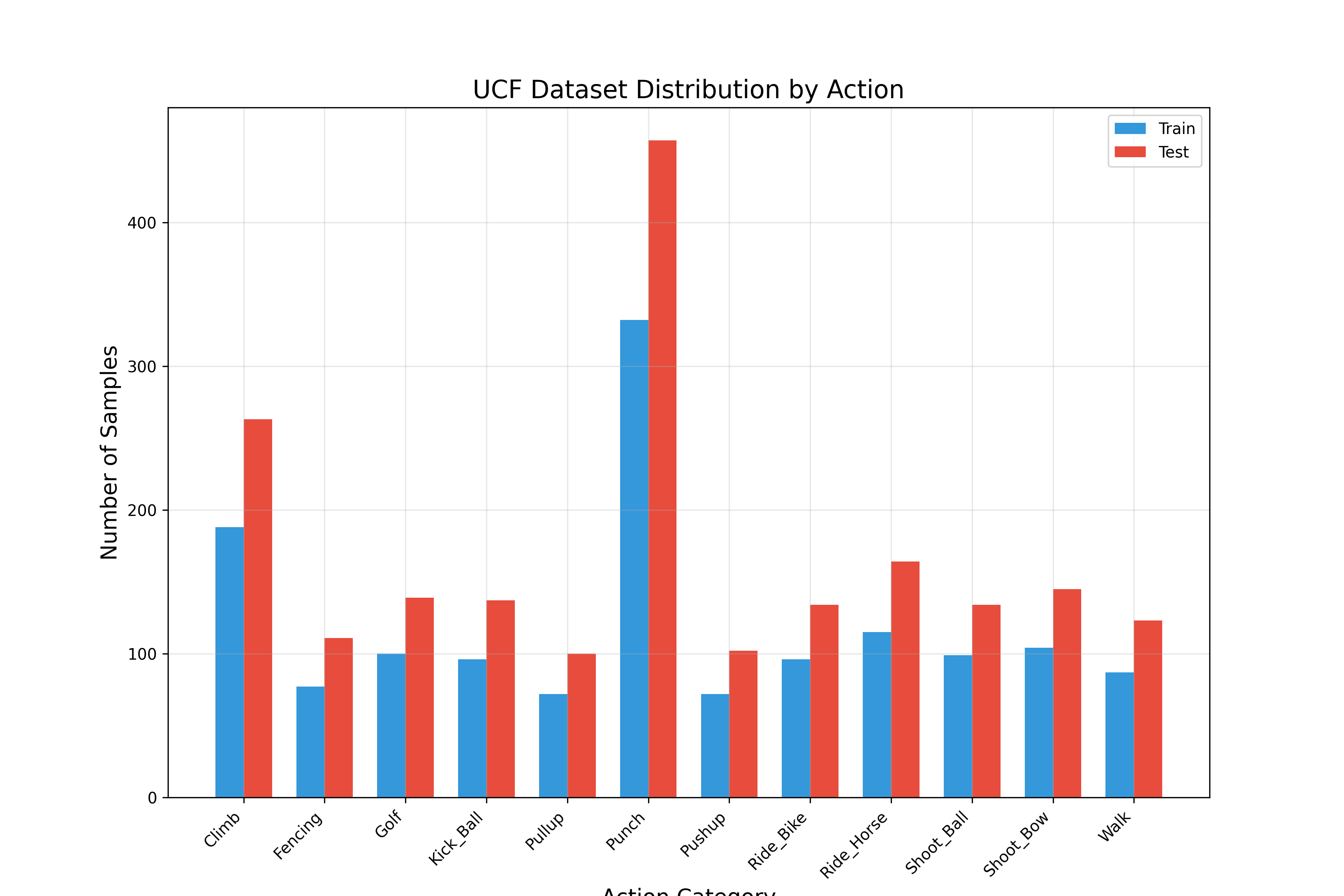}
    \caption{\textbf{UCF101 Distribution.}}
    \label{fig:HMDBSpeciesDistribution}
\end{figure}

\subsection{Self-Supervised Frozen Backbones}

\subsubsection{Stable Diffusion v2 Extraction}
\label{supp:sdv2}

Stable Video Diffusion builds upon Stable Diffusion v2 (SDv2) by adding temporal layers in between the layers of a pretrained SDv2, and training only them on video data. The output of these temporal layers is gated by a learned continuous $\alpha$ parameter between 0 and 1 for each layer. When $\alpha=1.0$, no information from the temporal is added and the model is effectively encoding the frames independently using SDv2. Therefore, we set $\alpha=1.0$ for all layers in order to use the same backbone to extract SDv2 features.

\subsubsection{V-JEPA and VideoMAEv2 Extraction}
\label{supp:videomae}
We extract VideoMAEv2 features using the official repository and pretrained encoder provided by the authors. Specifically, we compute spatiotemporal representations for each video segment and then average across the spatial dimension to obtain a single feature vector.

\subsection{Action Localization}
\label{supp:localization}

\begin{figure}[h]
    \centering
    \includegraphics[trim={3cm 0 7cm 0},clip,width=0.2\textwidth]{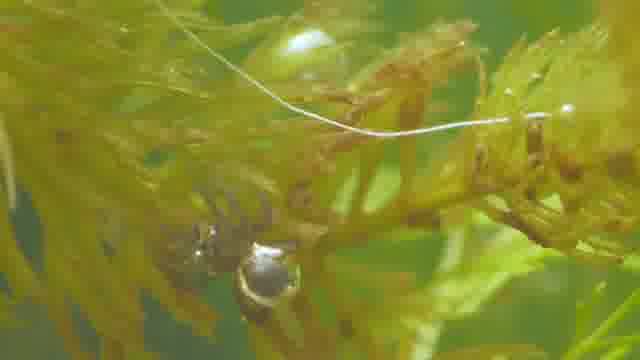}
    \includegraphics[trim={2cm 0 3.5cm 0},clip,width=0.2\textwidth]{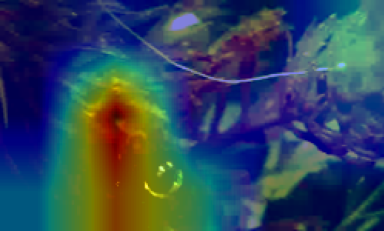}
    \caption{Action localization}
    \label{fig:heatmap}
\end{figure}

We can also analyze which parts of a video most contribute to the prediction of a specific action. For a given video, we can crop it into fixed patches and provide as input to the model the same patch over time. Then we can evaluate the prediction of an action over all patches and produce a heatmap like in Figure \ref{fig:heatmap}, which shows which parts of a video are most predicted as a prey being eaten, and we can localize the action in space. Resolution is not an issue for our model since we average pool the video features.

\newpage

\subsection{Cross-Species Generalization}
\label{supp:cross-species}

\begin{figure}[h]
    \centering
    \includegraphics[width=0.5\textwidth]{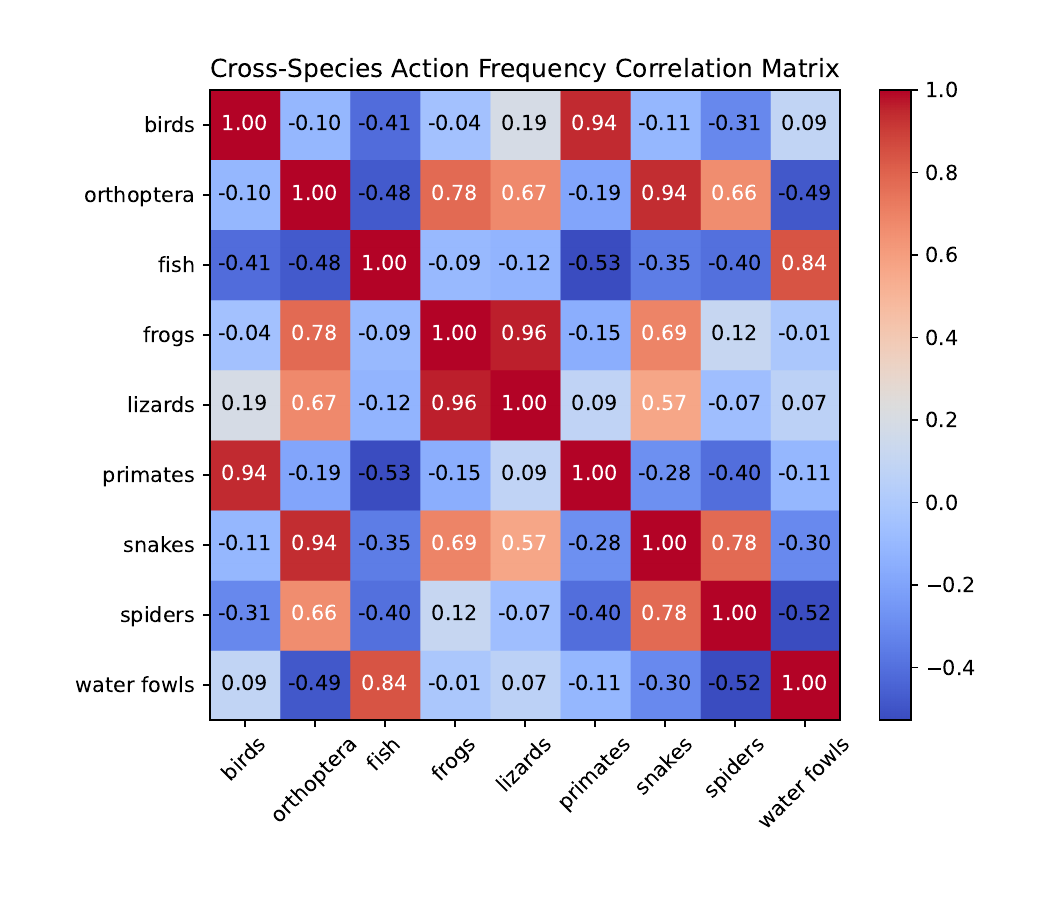}
    \caption{{\bf Action frequencies correlations across species.}}
    \label{fig:correlation}
\end{figure}

\begin{figure}[h]
    \centering
    \includegraphics[width=0.5\textwidth]{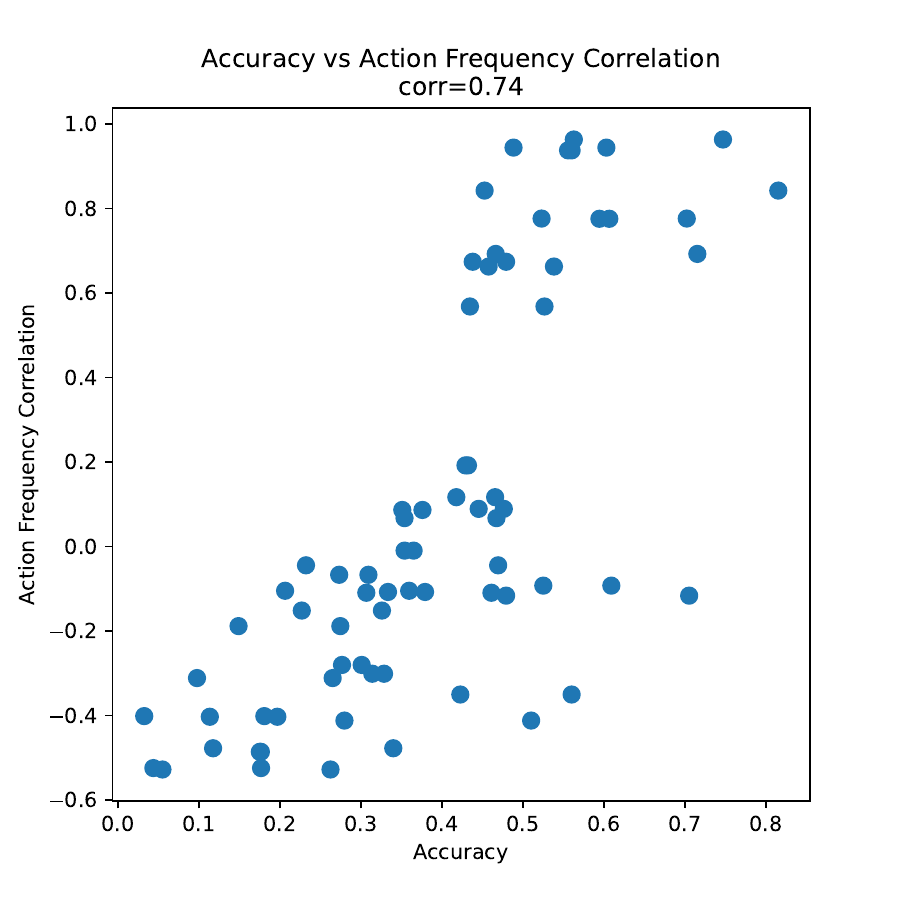}
    \caption{{\bf Accuracy vs action frequency correlation across species.}}
    \label{fig:acc_vs_corr}
\end{figure}

\begin{figure}[h]
    \centering
    \includegraphics[width=0.5\textwidth]{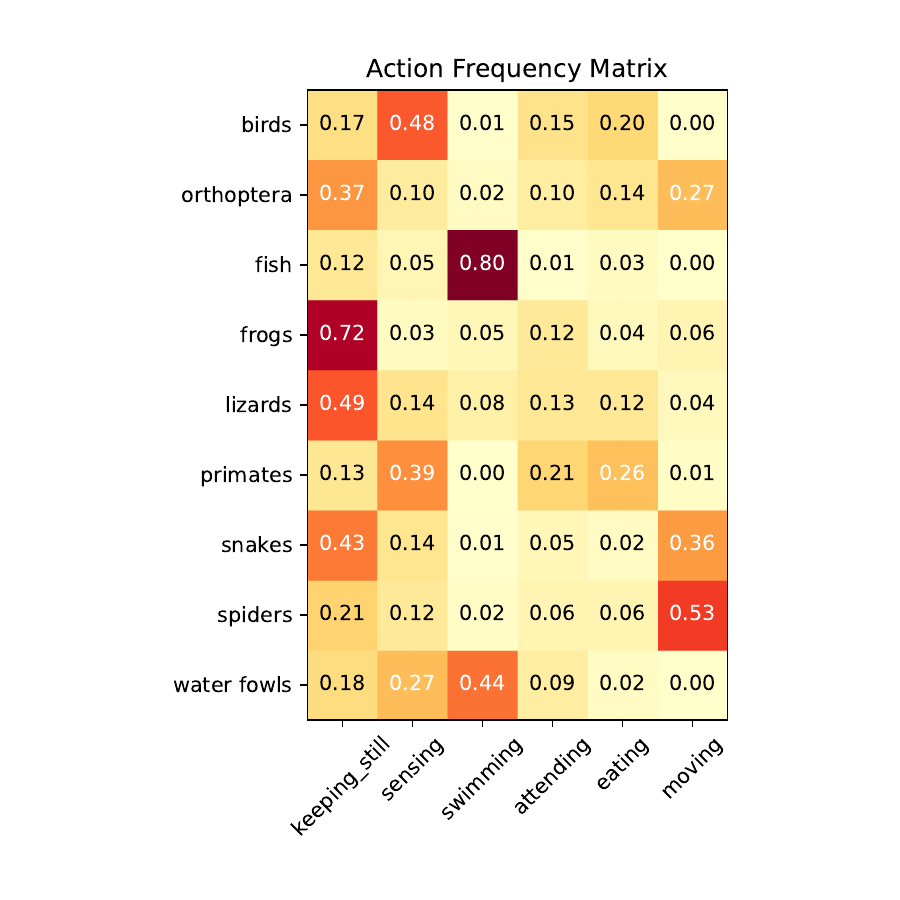}
    \caption{{\bf Action frequencies across species.}}
    \label{fig:freqs}
\end{figure}

\begin{figure}[h]
    \centering
    \includegraphics[width=0.5\textwidth]{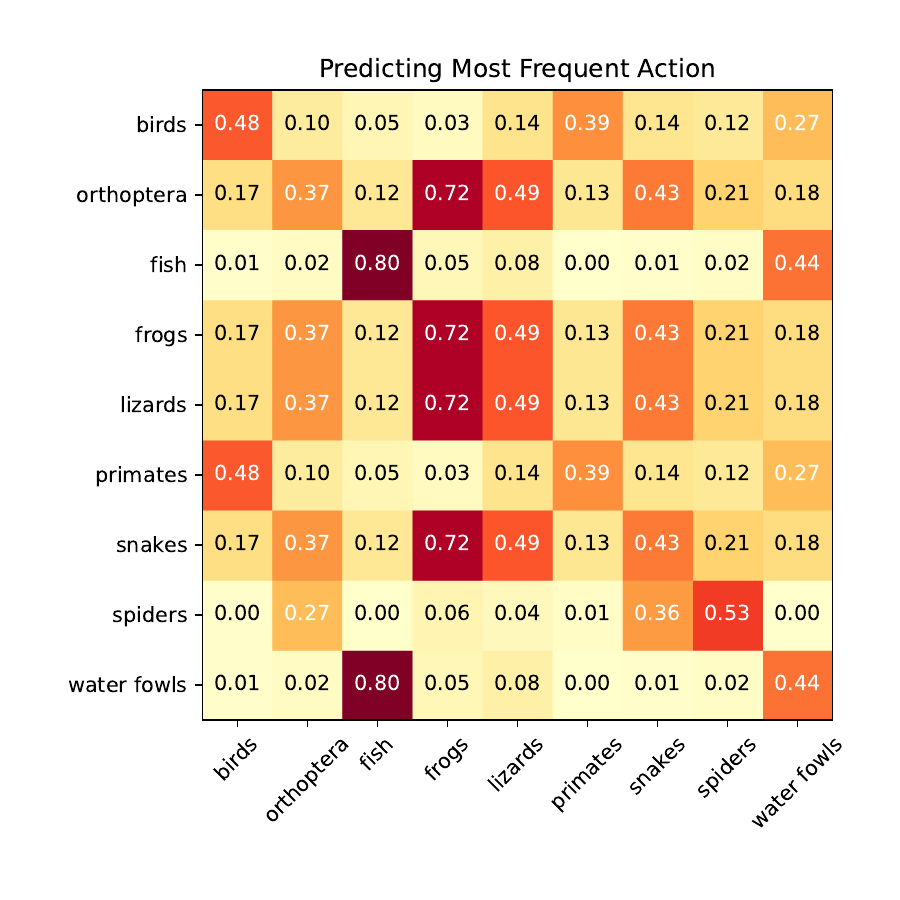}
    \caption{{\bf Results of predicting the most frequent action from the training species in the test species.}}
    \label{fig:bayesian}
\end{figure}

\begin{figure}[h]
    \centering
    \includegraphics[width=0.5\textwidth]{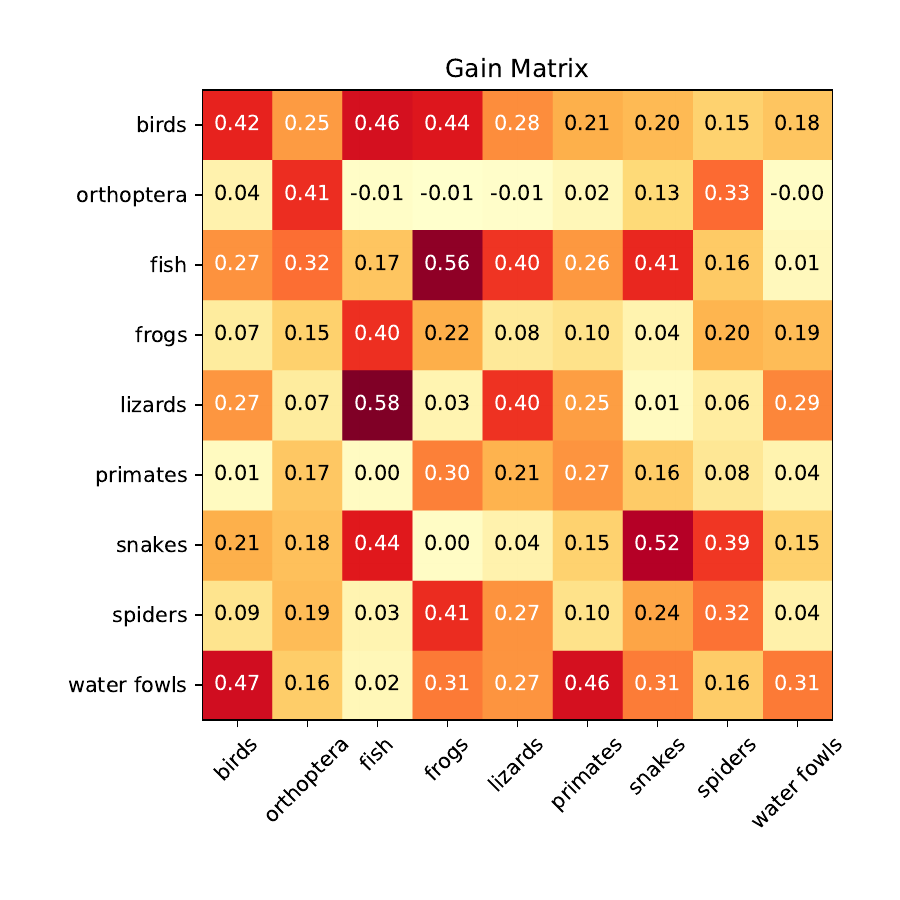}
    \caption{{\bf ActionDiff gains over predicting the most frequent class from the training species in the test species.}}
    \label{fig:gains}
\end{figure}

\newpage

\twocolumn[
\subsection{Choice of UNet Layer}
\label{sec:unet-layer}
\vspace{0.5cm}
\begin{center}
\begin{minipage}{\textwidth}
\centering
\begin{tabular}{lcccccc}
    \toprule
     & \multicolumn{2}{c}{Animal Kingdom} & \multicolumn{2}{c}{Charades-Ego} & \multicolumn{2}{c}{UCF101-HMDB51} \\
    \midrule
    \multirow{2}{*}{Layer} & Full           & Unseen  & $1^{st}\rightarrow1^{st}$ & $3^{rd}\rightarrow1^{st}$ & U $\rightarrow$ H & H $\rightarrow$ U \\
          & Dataset (mAP)  & Species (acc)  & mAP & mAP & Acc & Acc \\
    \midrule
    \multicolumn{3}{l}{\em ActionDiff (ours)} \\
    \midrule
    Layer 3 & 55.16 & 35.51 & 16.98 & 15.1 & 17.17 & 17.57 \\
    Layer 6 & 71.12 & 40.90 & 21.07 & 16.6 & 33.92 & 39.67 \\
    {\bf Layer 9} & {\bf 80.79} & {\bf 51.49} & \bf{36.5} & {\bf 30.2} & {\bf 75.6} & {\bf 81.5} \\
    \bottomrule
\end{tabular}
\captionof{table}{
{\bf Effect of UNet layer on all datasets.} Results improve with deeper decoder layers across all metrics. {\bf Bold} denotes best result.
}
\label{tab:layers}
\end{minipage}
\end{center}
\vspace{0.5cm}
]

\end{document}